\documentclass[11pt,a4paper]{article}

\usepackage[T1]{fontenc}

\usepackage{mdframed}
\usepackage{booktabs}
\usepackage{graphicx}

\usepackage[usenames,dvipsnames,svgnames]{xcolor}

\newcommand{\ie}{{\em i.e.}}
\newcommand{\eg}{{\em e.g.}}

\usepackage{amsmath,amsthm,amsfonts,amssymb,bm}

\DeclareMathOperator*{\argmax}{arg\,max} % thin space, limits on side in displays

\usepackage{framed}

\usepackage{varioref}

\usepackage{float}

\usepackage{multirow}
\usepackage[hyperref]{acl2017}
\usepackage{times}
\usepackage{latexsym}

\usepackage{url}

\aclfinalcopy % Uncomment this line for the final submission
\def\aclpaperid{164} %  Enter the acl Paper ID here

\title{Learning to Ask: Neural Question Generation for \\
       Reading Comprehension}
\small
\author{Xinya Du{\normalfont \textsuperscript{1}}  \quad \ \ Junru Shao{\normalfont \textsuperscript{2}} \quad \ \ Claire Cardie{\normalfont \textsuperscript{1}}\\
  \textsuperscript{1}Department of Computer Science, Cornell University\\
   \textsuperscript{2}Zhiyuan College, Shanghai Jiao Tong University\\
  {\tt \{xdu, cardie\}@cs.cornell.edu \  yz\_sjr@sjtu.edu.cn  } \\
  }
\normalsize

\date{}

\begin{document}
\maketitle
\begin{abstract}

We study automatic question generation for sentences from text passages in reading comprehension. We introduce an attention-based sequence learning model for the task and investigate the effect of encoding sentence- vs.\ paragraph-level information. In contrast to all previous work, our model does not rely on hand-crafted rules or a sophisticated NLP pipeline;  it is instead trainable end-to-end via sequence-to-sequence learning. Automatic evaluation results show that our system significantly outperforms the state-of-the-art rule-based system. In human evaluations, questions generated by our system are also rated as being more natural (\ie, grammaticality, fluency) and as more difficult to answer (in terms of syntactic and lexical divergence from the original text and reasoning needed to answer).
\end{abstract}

\section{Introduction}

% Introduction Provide Motivation, explain why the what you are doing is important. Introduction Provide Motivation, explain why the what you are doing is important.

Question generation (QG) aims to create natural questions from a given a sentence or paragraph.
%QG is an important measure of the system's ability to extract useful information in the text, and the understanding of the text. 
%
One key application of question generation is in the area of education --- to generate questions for reading comprehension materials~\cite{heilman2010good}. Figure~\ref{fig:problem}, for example, shows three manually generated questions that test a user's understanding of the associated text passage.
Question generation systems can also be deployed
as chatbot components (\eg, asking questions to start a conversation or to request feedback~\cite{nasrin2016vqg}) or, arguably,
%in human-robot communication~\cite{surveyhumanrobot03}; [refs]; 
as a clinical tool for evaluating or improving mental health~\cite{Weizenbaum1966eliza, parry1971colby}.

In addition to the above applications, question generation systems can aid in the development of annotated data sets for natural language processing (NLP) research in reading comprehension and question answering. Indeed the creation of such datasets, \eg, SQuAD~\cite{rajpurkar2016squad} and MS MARCO~\cite{nguyen2016msmarco}, has spurred research in these areas.

For the most part, question generation has been tackled in the past via rule-based approaches (\eg,~\newcite{mitkov2003computer, rus2010first}. The success of these approaches hinges critically on the existence of well-designed rules for declarative-to-interrogative sentence transformation, typically based on deep linguistic knowledge. 

\begin{figure}[t]
\centering
%\toprule[0.05cm]
\begin{tabular}{p{6.5cm}} 
\toprule
    \small \textbf{Sentence}:\\
    {
    \small
    \fontfamily{phv} \selectfont 
    Oxygen is used in cellular respiration and released by \textcolor{LimeGreen}{photosynthesis}, which uses the energy of \textcolor{red}{sunlight} to produce oxygen from \textcolor{blue}{water}. 
    }\\[4pt]
    \small \textbf{Questions}: \\
    
     % 1st question
    {
    \small \fontfamily{phv}\selectfont 
    -- What life process produces oxygen in the presence of light?
    }\\
    {
    \small \fontfamily{phv}\selectfont 
    \quad \emph{\textcolor{LimeGreen}{photosynthesis}}
    }\\[4pt]
    
    % 2nd question
    {
    \small
    \fontfamily{phv}\selectfont 
    -- Photosynthesis uses which energy to form oxygen from water? 
    }\\
    {
    \small
    \fontfamily{phv}\selectfont 
    \quad \emph{\textcolor{red}{sunlight}}
    }\\[4pt]
    
    % 3rd question
    {
    \small \fontfamily{phv}\selectfont 
    -- From what does photosynthesis get oxygen?
    }\\
    {
    \small \fontfamily{phv}\selectfont 
    \quad \emph{\textcolor{Blue}{water}}
    }\\[2pt]
\bottomrule
\end{tabular}
%\bottomrule[0.05cm]
\caption{Sample sentence from the second paragraph of the article \textit{Oxygen}, along with the natural questions and their answers. \label{fig:problem}}
\vspace{-0.2cm}
\end{figure}

%~\cite{chomsky1971conditions} 
To improve over a purely rule-based system, \newcite{heilman2010good} introduced an overgenerate-and-rank approach that generates multiple questions from an input sentence using a rule-based approach and then ranks them using a supervised learning-based ranker.  Although the ranking algorithm helps to produce more acceptable questions, it relies heavily on a manually crafted feature set, and the questions generated often overlap word for word with the tokens in the input sentence, making them very easy to answer.

% human evaluations, not realistic for development of learning-based systems.

\newcite{vanderwende2008importance} point out that learning to ask good questions is an important task in NLP research in its own right, and should consist of more than the syntactic transformation of a declarative sentence.
In particular, a natural sounding question often compresses the sentence
on which it is based (\eg,~question 3 in Figure~\ref{fig:problem}), uses synonyms for terms in the passage (\eg,~``form'' for ``produce'' in question 2 and ``get'' for ``produce'' in question 3), or refers to entities from preceding sentences or clauses (\eg,~the use of ``photosynthesis'' in question 2).
Othertimes, world knowledge is employed to produce a good question
(\eg,~identifying ``photosynthesis'' as a ``life process'' in question 1).
%In order to ask the questions, one has to compress/summarize the sentence snippet, recognize the relationship between different entities. Sometimes world knowledge is required (\eg, identifying photosynthesis as a life process in the third question). 
%
In short, constructing natural questions of reasonable difficulty would seem to require an~\emph{abstractive} approach that can produce fluent phrasings that do not exactly match the text from which they were drawn.

As a result, and in contrast to all previous work, we propose here to
frame the task of question generation as a
%end-to-end 
sequence-to-sequence learning problem that directly maps a sentence  from a text passage to a question. Importantly, our approach is
fully data-driven in that it requires no manually generated rules.  

More specifically, inspired by the recent success in neural machine translation~\cite{sutskever2014sequence, bahdanau2014neural}, summarization~\cite{rush2015namas, iyer2016code}, and image caption generation~\cite{xu2015show}, we tackle question generation using a conditional neural language model with a global attention mechanism~\cite{luong2015effective}. 
%To encode the sentence snippet, we investigate several models, .
We investigate several variations of this model, including one that takes into account paragraph- rather than sentence-level information from the reading passage as well as other variations that determine the importance of pre-trained vs.\ learned word embeddings. 

In evaluations on the SQuAD dataset~\cite{rajpurkar2016squad} using three automatic 
evaluation metrics,  we find that our system 
% statistically 
significantly
outperforms a collection of strong baselines, including an information retrieval-based system~\cite{robertson1994bm25}, 
%ctc: readers won't know what this is 
%extraction from input sentence, 
a statistical machine translation approach~\cite{koehn2007moses}, and the overgenerate-and-rank  approach of ~\newcite{heilman2010good}. 
%For example, our best system achieves BLEU score of 12.28, and a ROUGE\textsubscript{L} score of 39.75, compared to that of H\&S system, 11.18 (BLEU) and 30.98 (ROUGE\textsubscript{L}). 
Human evaluations also rated our generated questions as more grammatical, fluent, and challenging (in terms of syntactic divergence from the original reading passage and reasoning needed to answer) than the state-of-the-art ~\newcite{heilman2010good} system.

In the sections below we discuss related work (Section~\ref{sec:related}), specify the task definition (Section~\ref{sec:task}) and describe our neural sequence learning based models (Section~\ref{sec:model}). We explain the experimental setup in Section~\ref{sec:experiment}. Lastly, we present the evaluation results as well as a detailed analysis.

\section{Related Work}
\label{sec:related}

\textbf{Reading Comprehension} is a challenging task for machines, requiring both understanding of natural language and knowledge of the world~\cite{rajpurkar2016squad}. Recently many new datasets have been released and in most of these datasets, the questions are generated in a synthetic way. For example, bAbI~\cite{weston2015towards} is a fully synthetic dataset featuring 20 different tasks.~\newcite{hermann2015teaching} released a corpus of cloze style questions by replacing entities with placeholders in abstractive summaries of CNN/Daily Mail news articles.~\newcite{danqi2016exam} claim that the CNN/Daily Mail dataset is easier than previously thought, and their system almost reaches the ceiling performance.~\newcite{richardson2013mctest} curated MCTest, in which crowdworker questions are paired with four answer choices. Although MCTest contains challenging natural questions, it is too small for training data-demanding question answering models.

Recently,~\newcite{rajpurkar2016squad} released the Stanford Question Answering Dataset\footnote{\url{https://stanford-qa.com}} (SQuAD), which overcomes the aforementioned small size and (semi-)synthetic issues. The questions are posed by crowd workers and are of relatively high quality. We use SQuAD in our work, and similarly, we focus on the generation of natural questions for reading comprehension materials, albeit via automatic means.

\textbf{Question Generation} has attracted the attention of the natural language generation (NLG) community in recent years, since the work of~\newcite{rus2010first}.

Most work tackles the task with a rule-based approach. Generally, they first transform the input sentence into its syntactic representation, which they then use to generate an interrogative sentence. A lot of research has focused on first manually constructing question templates, and then applying them to generate questions~\cite{mostow2009generating, lindberg2013online, mazidi2014linguistic}.~\newcite{labutov2015deep} use crowdsourcing to collect a set of templates and then rank the relevant templates for the text of another domain. Generally, the rule-based approaches make use of the syntactic roles of words, but not their semantic roles. 

~\newcite{heilman2010good} introduce an overgenerate-and-rank approach: their system first overgenerates questions and then ranks them. Although they incorporate learning to rank, their system's performance still depends critically on the manually constructed generating rules.~\newcite{nasrin2016vqg} introduce visual question generation task, to explore the deep connection between language and vision.~\newcite{serban2016factoid30m} propose generating simple factoid questions from logic triple (subject, relation, object). 
% Compared with our task, 
Their task tackles mapping from structured representation to natural language text, and their generated questions are consistent in terms of format and diverge much less than ours. 

To our knowledge, none of the previous works has framed QG for reading comprehension in an end-to-end fashion, and nor have them used deep sequence-to-sequence learning approach to generate questions.

\section{Task Definition}
\label{sec:task}

In this section, we define the question generation task. Given an input sentence $\mathbf{x}$, our goal is to generate a natural question $\mathbf{y}$ related to information in the sentence, $\mathbf{y}$ can be a sequence of an arbitrary length: $[ y_1, ..., y_{|\mathbf{y}|}]$. Suppose the length of the input sentence is $M$, $\mathbf{x}$ could then be represented as a sequence of tokens~$[ x_1, ..., x_M]$. The QG task is defined as finding $\mathbf{\overline{y}}$, such that:

\begin{equation}
\label{equ:task}
\mathbf{\overline{y}} = \argmax_{\mathbf{y}} P \left( \mathbf{y} \vert \mathbf{x} \right)
\end{equation}
where $P \left(\mathbf{y} \vert \mathbf{x} \right)$ is the conditional log-likelihood of the predicted question sequence $\mathbf{y}$, given the input $\mathbf{x}$. In section~\ref{ssec:decoder}, we will elaborate on the global attention mechanism for modeling $P \left(\mathbf{y} \vert \mathbf{x} \right)$.

\section{Model}
\label{sec:model}

Our model is partially inspired by the way in which a human would solve the task. To ask a natural question, people usually pay attention to certain parts of the input sentence, as well as associating context information from the paragraph. We model the conditional probability using RNN encoder-decoder architecture~\cite{bahdanau2014neural, cho2014phrase}, and adopt the global attention mechanism~\cite{luong2015effective} to make the model focus on certain elements of the input when generating each word during decoding. 

Here, we investigate two variations of our models: one that only encodes the sentence and another that encodes both sentence and paragraph-level information. 

\subsection{Decoder}
\label{ssec:decoder}
Similar to~\newcite{sutskever2014sequence} and~\newcite{chopra2016abstractive}, we factorize the the conditional in equation~\ref{equ:task} into a product of word-level predictions:
\begin{equation}
\nonumber
P \left( \mathbf{y} \vert \mathbf{x} \right) = \prod_{t=1}^{\vert y \vert} P \left( y_{t} \vert \mathbf{x}, y_{ <t}\right)
\end{equation}
where probability of each $y_t$ is predicted based on all the words that are generated previously (\ie, $y_{ <t}$), and input sentence $\mathbf{x}$.

More specifically, 
\begin{equation}
P \left( y_{t} \vert \mathbf{x}, y_{ <t}\right) = \textnormal{softmax} \left(\mathbf{W}_s \textnormal{tanh}\left( \mathbf{W}_t [ \mathbf{h}_t; \mathbf{c}_t]  \right )\right)
\end{equation}
with $\mathbf{h}_t$ being the recurrent neural networks state variable at time step $t$, and $\mathbf{c}_t$ being the attention-based encoding of $\mathbf{x}$ at decoding time step $t$ (Section~\ref{ssec:encoder}). $\mathbf{W}_s$ and $\mathbf{W}_t$ are parameters to be learned.

\begin{equation}
\mathbf{h}_t = \textnormal{LSTM}_1 \left( y_{t-1} , \mathbf{h}_{t-1}\right) 
\end{equation}
here, $\textnormal{LSTM}$ is the Long Short-Term Memory (LSTM) network~\cite{hochreiter1997long}. It generates the new state $\mathbf{h}_t$, given the representation of previously generated word $y_{t-1}$ (obtained from a word look-up table), and the previous state $\mathbf{h}_{t-1}$.

The initialization of the decoder's hidden state differentiates our basic model and the model that incorporates paragraph-level information.

For the basic model, it is initialized by the sentence's representation $\mathbf{s}$ obtained from the sentence encoder (Section~\ref{ssec:encoder}). For our paragraph-level model, the \emph{concatenation} of the sentence encoder's output $\mathbf{s}$ and the paragraph encoder's output $\mathbf{s}'$ is used as the initialization of decoder hidden state. To be more specific, the architecture of our paragraph-level model is like a ``Y''-shaped network which encodes both sentence- and paragraph-level information via two RNN branches and uses the concatenated representation for decoding the questions.

\subsection{Encoder}
\label{ssec:encoder}

% We next describe our encoder models. 
The attention-based sentence encoder is used in both of our models, while the paragraph encoder is only used in the model that incorporates paragraph-level information.

\vspace{0.3cm}

\noindent \textbf{Attention-based sentence encoder}:

\vspace{0.01cm}

We use a bidirectional LSTM to encode the sentence,
\begin{equation}
\nonumber
\begin{split}
\overrightarrow{\mathbf{b}_t} &= \overrightarrow{{\textnormal{LSTM}_2}} \left( x_{t}, \overrightarrow{\mathbf{b}_{t-1}}\right) \\
\overleftarrow{\mathbf{b}_t} &= \overleftarrow{{\textnormal{LSTM}_2}} \left( x_{t}, \overleftarrow{\mathbf{b}_{t+1}}\right) 
\end{split}
\end{equation}

where $\overrightarrow{\mathbf{b}_t}$ is the hidden state at time step $t$ for the forward pass LSTM, $\overleftarrow{\mathbf{b}_t}$ for the backward pass.

To get \emph{attention-based} encoding of $\mathbf{x}$ at decoding time step $t$, namely, $\mathbf{c}_t$, we first get the context dependent token representation by $\mathbf{b}_t = [\overrightarrow{\mathbf{b}_t}; \overleftarrow{\mathbf{b}_t} ]$, then we take the weighted average over $\mathbf{b}_t$ ($t = 1, ..., |\mathbf{x}|)$,

\begin{equation}
\mathbf{c}_t = \sum_{i=1,..,|\mathbf{x}|} a_{i,t} \mathbf{b}_i
\end{equation}

The attention weight are calculated by the bi-linear scoring function and softmax normalization,

\begin{equation}
a_{i,t} = \frac{\exp \left( \mathbf{h}_t^{T} \mathbf{W}_b \mathbf{b}_i\right)}{\sum_{j} \exp \left( \mathbf{h}_t^{T} \mathbf{W}_b \mathbf{b}_j\right)}  
\end{equation}

To get the sentence encoder's output for initialization of decoder hidden state, we concatenate last hidden state of the forward and backward pass, namely, $\mathbf{s} = [\overrightarrow{\mathbf{b}_{|\mathbf{x}|}}; \overleftarrow{\mathbf{b}_1} ]$. 

% define function $g$

\vspace{0.3cm}
\noindent \textbf{Paragraph encoder}:
\vspace{0.01cm}

Given sentence $\mathbf{x}$, we want to encode the paragraph containing $\mathbf{x}$. Since in practice the paragraph is very long, we set a length threshold $L$, and truncate the paragraph at the $L$\textsuperscript{th} token. We call the truncated paragraph ``paragraph'' henceforth. 

Denoting the paragraph as $\mathbf{z}$, we use another bidirectional LSTM to encode $\mathbf{z}$,
\begin{equation}
\nonumber
\begin{split}
\overrightarrow{\mathbf{d}_t} &= \overrightarrow{{\textnormal{LSTM}_3}} \left( z_{t}, \overrightarrow{\mathbf{d}_{t-1}}\right) \\
\overleftarrow{\mathbf{d}_t} &= \overleftarrow{{\textnormal{LSTM}_3}} \left( z_{t}, \overleftarrow{\mathbf{d}_{t+1}}\right) \\
% \mathbf{d}_t &= [ \overrightarrow{\mathbf{d}_t}, \overleftarrow{\mathbf{d}_t} ]
\end{split}
\end{equation}

With the last hidden state of the forward and backward pass, we use the concatenation $[\overrightarrow{\mathbf{d}_{|\mathbf{z}|}}; \overleftarrow{\mathbf{d}_1} ]$ as the paragraph encoder's output $\mathbf{s}'$.

\subsection{Training and Inference}

Giving a training corpus of sentence-question pairs: $\mathcal{S} = \left\{ \left( \mathbf{x}^{(i)}, \mathbf{y}^{(i)}\right) \right\}_{i=1}^{S}$, our models' training objective is to minimize the negative log-likelihood of the training data with respect to all the parameters, as denoted by $\theta$,

\begin{equation} \label{equ:loss}
\nonumber
\begin{split}
\mathcal{L} & = - \sum_{i=1}^{S} \log P \left( \mathbf{y}^{(i)}\vert \mathbf{x}^{(i)}; \theta \right) \\
 & = - \sum_{i=1}^{S} \sum_{j=1}^{\vert \mathbf{y}^{(i)} \vert} \log P \left( y_{j}^{(i)}\vert \mathbf{x}^{(i)}, y_{ <j}^{(i)} ; \theta \right)
\end{split}
\end{equation}

% \begin{equation}
% \mathbf{y} = \argmax_{y} \log P \left( y \vert \mathbf{x} \right)
% \end{equation}

Once the model is trained, we do inference using beam search. The beam search is parametrized by the possible paths number $k$. 

As there could be many rare words in the input sentence that are not in the target side dictionary, during decoding many \verb|UNK| tokens will be output. Thus, post-processing with the replacement of \verb|UNK| is necessary. Unlike~\newcite{luong2015addressing}, we use a simpler replacing strategy for our task. For the decoded \verb|UNK| token at time step $t$, we replace it with the token in the input sentence with the highest attention score, the index of which is $\argmax_{i} a_{i,t}$.

% \begin{equation} 
% \nonumber
% y_t = x{\argmax_{i} a_{i,t}
% \end{equation}

\section{Experimental Setup}
\label{sec:experiment}

We experiment with our neural question generation model on the processed SQuAD dataset. In this section, we firstly describe the corpus of the task. We then give implementation details of our neural generation model, the baselines to compare, and their experimental settings. Lastly, we introduce the evaluation methods by automatic metrics and human raters.

\subsection{Dataset} 
\label{ssec:dataset}

\begin{figure}[t]
\centering
\includegraphics[scale=.52]{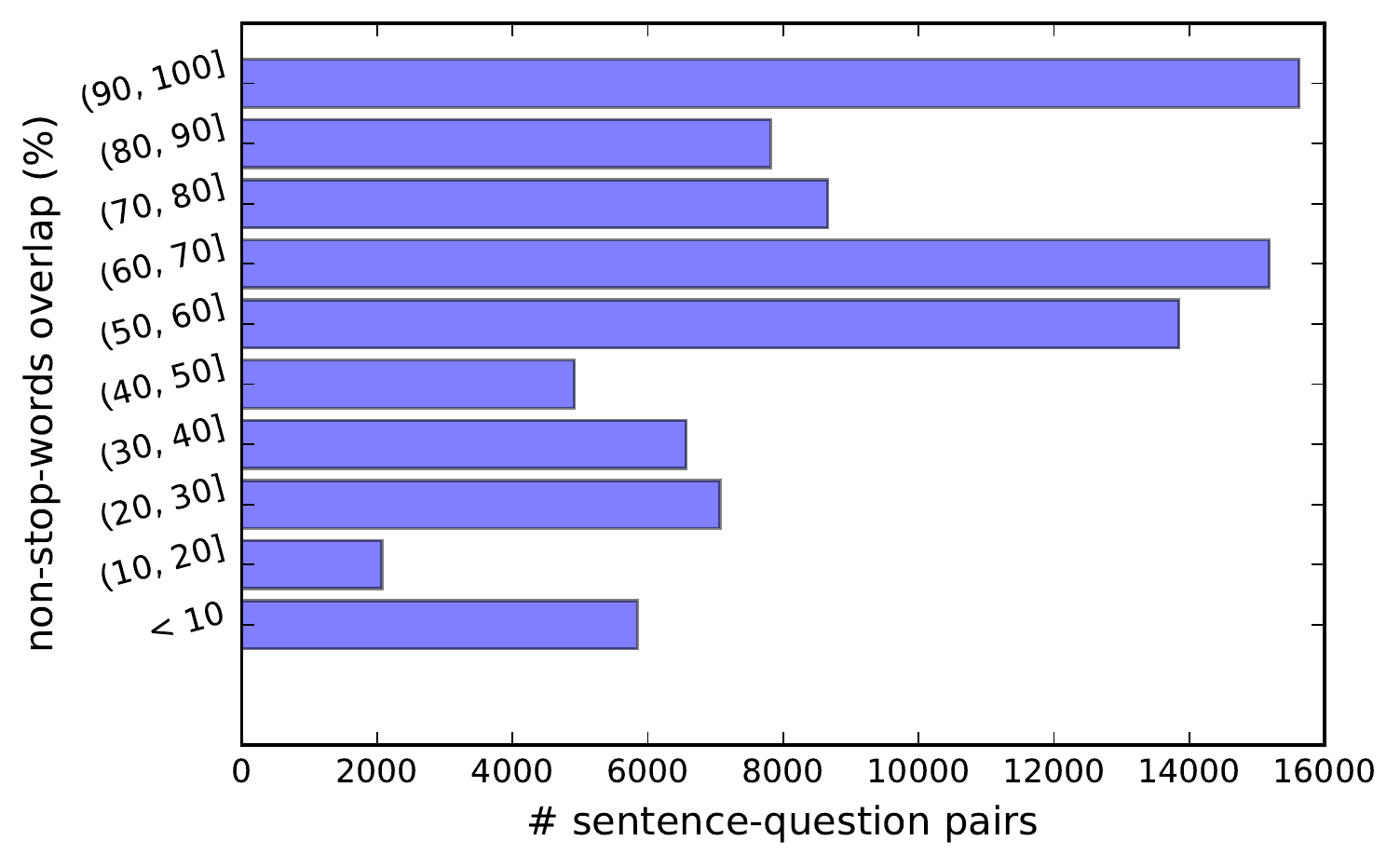}
\vspace{-0.4cm}
\caption{Overlap percentage of sentence-question pairs in training set. $y$-axis is \# non-stop-words overlap with respect to the total \# tokens in the question (a percentage); $x$-axis is \# sentence-question pairs for a given overlap percentage range.}
\vspace{-0.4cm}
\label{fig:overlap}
\end{figure}

With the SQuAD dataset~\cite{rajpurkar2016squad}, we extract sentences and pair them with the questions. We train our models with the sentence-question pairs. The dataset contains 536 articles with over 100k questions posed about the articles. The authors employ Amazon Mechanical Turks crowd-workers to create questions based on the Wikipedia articles. Workers are encouraged to use their own words without any copying phrases from the paragraph. Later, other crowd-workers are employed to provide answers to the questions. The answers are spans of tokens in the passage.

Since there is a hidden part of the original SQuAD that we do not have access to, we treat the accessible parts ($\sim$90\%) as the entire dataset henceforth. 

 We first run Stanford CoreNLP~\cite{manning2014corenlp} for pre-processing: tokenization and sentence splitting. We then lower-case the entire dataset. With the offset of the answer to each question, we locate the sentence containing the answer and use it as the input sentence. In some cases (<~0.17\% in training set), the answer spans two or more sentences, and we then use the concatenation of the sentences as the input ``sentence''.

 Figure~\ref{fig:overlap} shows the distribution of the token overlap percentage of the sentence-question pairs. Although most of the pairs have over 50\% overlap rate, about 6.67\% of the pairs have no non-stop-words in common, and this is mostly because of the answer offset error introduced during annotation. Therefore, we prune the training set based on the constraint: the sentence-question pair must have at least one non-stop-word in common. Lastly we add \verb|<SOS>| to the beginning of the sentences, and \verb|<EOS>| to the end of them.

We randomly divide the dataset at the article-level into a training set (80\%), a development set (10\%), and a test set (10\%). We report results on the 10\% test set.

Table~\ref{tab:data} provides some statistics on the processed dataset: there are around 70k training samples, the sentences are around 30 tokens, and the questions are around 10 tokens on average. For each sentence, there might be multiple corresponding questions, and, on average, there are 1.4 questions for each sentence.

\begin{table}[t]
\centering

    \begin{tabular}{lc} 
    \toprule
    % \bf{Model} & \bf{BLEU 1}  & \bf{BLEU 2} & \bf{BLEU 3} & \textbf{BLEU 4} & \bf{Rouge-L} & \textbf{Meteor}\\\midrule
    %   Item & \\\midrule
    \# pairs (Train) & 70484\\
    \# pairs (Dev) & 10570\\
     \# pairs (Test) & 11877\\ \cmidrule(r{2pt}){1-2}
     Sentence: avg. tokens & 32.9\\
     Question: avg. tokens & 11.3\\\cmidrule(r{2pt}){1-2}
     Avg. \# questions per sentence& 1.4\\
    \bottomrule
  
    \end{tabular}
  \caption{Dataset (processed) statistics. Sentence average \# tokens, question average \# tokens, and average \# questions per sentence statistics are from training set. These averages are close to the statistics on development set and test set.}
\label{tab:data}
\vspace{-0.2cm}
\end{table}

\subsection{Implementation Details}

\begin{table*}[htb]
\centering
\resizebox{\textwidth}{!}{%
    \begin{tabular}{l|cccc|c|c} 
    \toprule
      Model & BLEU 1  & BLEU 2 & BLEU 3 & BLEU 4 & METEOR & ROUGE\textsubscript{L}\\\midrule
      
IR\textsubscript{BM25}          & 5.18  & 0.91  & 0.28  & 0.12  & 4.57  & 9.16  \\
IR\textsubscript{Edit Distance} & 18.28 & 5.48  & 2.26  & 1.06  & 7.73  & 20.77 \\
MOSES+                            & 15.61 & 3.64  & 1.00  & 0.30  & 10.47 & 17.82 \\
DirectIn                          & 31.71 & 21.18 & 15.11 & 11.20 & 14.95 & 22.47 \\
H\&S                              & 38.50 & 22.80 & 15.52 & 11.18 & 15.95 & 30.98 \\
Vanilla seq2seq                     & 31.34 & 13.79 & 7.36  & 4.26  & 9.88  & 29.75 \\\midrule
Our model (no pre-trained)           & 41.00 & 23.78 & 15.71 & 10.80 & 15.17 & 37.95 \\
Our model (w/ pre-trained)           & \textbf{43.09} & \textbf{25.96} & \textbf{17.50} & \textbf{12.28} & \textbf{16.62} & \textbf{39.75} \\
% \quad + paragraph                & 42.54 & 25.33 & 16.98 & 11.86 & 16.28 & 39.37 \\
\quad + paragraph               & 42.54 & 25.33 & 16.98 & 11.86 & 16.28 & 39.37\\
% \quad + title &43.52&	25.16 &	16.87	&11.74	&16.13&	39.49 \\
    \bottomrule 
\end{tabular}}
  \caption{Automatic evaluation results of different systems by BLEU 1--4, METEOR and ROUGE\textsubscript{L}. For a detailed explanation of the baseline systems, please refer to Section~\ref{ssec:baselines}. The best performing system for each column is highlighted in boldface. Our system which encodes only sentence with pre-trained word embeddings achieves the best performance across all the metrics.}
 
\label{tab:results}
\end{table*}

We implement our models~\footnote{The code is available at~\url{https://github.com/xinyadu/nqg}.} in Torch7~\footnote{\url{http://torch.ch/}} on top of the newly released~OpenNMT system~\cite{2017opennmt}.

For the source side vocabulary $\mathcal{V}$, we only keep the 45k most frequent tokens (including \verb|<SOS>|, \verb|<EOS>| and placeholders). For the target side vocabulary $\mathcal{U}$, similarly, we keep the 28k most frequent tokens. All other tokens outside the vocabulary list are replaced by the \verb|UNK| symbol. We choose word embedding of 300 dimensions and use the \verb|glove.840B.300d| pre-trained embeddings~\cite{pennington2014glove} for initialization. We fix the word representations during training.

We set the LSTM hidden unit size to 600 and set the number of layers of LSTMs to 2 in both the encoder and the decoder. Optimization is performed using stochastic gradient descent (SGD), with an initial learning rate of 1.0. We start halving the learning rate at epoch 8. The mini-batch size for the update is set at 64. Dropout with probability 0.3 is applied between vertical LSTM stacks. We clip the gradient when the its norm exceeds 5.

All our models are trained on a single GPU. We run the training for up to 15 epochs, which takes approximately 2 hours. We select the model that achieves the lowest perplexity on the dev set.

During decoding, we do beam search with a beam size of 3. Decoding stops when every beam in the stack generates the \verb|<EOS>| token.

% \penalty -15000

All hyperparameters of our model are tuned using the development set. The results are reported on the test set.

\subsection{Baselines}
\label{ssec:baselines}

To prove the effectiveness of our system, we compare it to several competitive systems. Next, we briefly introduce their approaches and the experimental setting to run them for our problem. Their results are shown in Table~\ref{tab:results}.

\vspace{0.1cm} \noindent \textbf{IR} stands for our information retrieval baselines. Similar to ~\newcite{rush2015namas}, we implement the IR baselines to control memorizing questions from the training set. We use two metrics to calculate the distance between a question and the input sentence, \ie, BM-25~\cite{robertson1994bm25} and edit distance~\cite{levenshtein1966binary}. According to the metric, the system retrieves the training set to find the question with the highest score.

\vspace{0.1cm} \noindent \textbf{MOSES+}~\cite{koehn2007moses} is a widely used phrase-based statistical machine translation system. Here, we treat sentences as source language text, we treat questions as target language text, and we perform the translation from sentences to questions. We train a tri-gram language model on target side texts with KenLM~\cite{KenLM2015}, and tune the system with MERT on dev set. Performance results are reported on the test set.

\vspace{0.1cm} \noindent \textbf{DirectIn} is an intuitive yet meaningful baseline in which the longest sub-sentence of the sentence is \emph{directly} taken as the predicted question.~\footnote{We also tried using the \emph{entire} input sentence as the prediction output, but the performance is worse than taking sub-sentence as the prediction, across all the automatic metrics except for METEOR.} To split the sentence into sub-sentences, we use a set of splitters, \ie, \{``?'', ``!'', ``,'', ``.'', ``;''\}.

\vspace{0.1cm} \noindent \textbf{H\&S} is the rule-based overgenerate-and-rank system that was mentioned in Section~\ref{sec:related}. When running the system, we set the parameter \verb|just-wh| true (to restrict the output of the system to being only wh-questions) and set \verb|max-length| equal to the longest sentence in the training set. We also set \verb|downweight-pro| true, to down weight questions with unresolved pronouns so that they appear towards the end of the ranked list. For comparison with our systems, we take the top question in the ranked list.

\vspace{0.1cm} \noindent \textbf{Seq2seq}~\cite{sutskever2014sequence} is a basic encoder-decoder sequence learning system for machine translation. We implement their model in Tensorflow. The input sequence is reversed before training or translating. Hyperparameters are tuned with dev set. We select the model with the lowest perplexity on the dev set.

\subsection{Automatic Evaluation} 

We use the evaluation package released by~\newcite{chen2015microsoft}, which was originally used to score image captions. The package includes BLEU 1, BLEU 2, BLEU 3, BLEU 4~\cite{papineni2002bleu}, METEOR~\cite{2014meteor} and ROUGE\textsubscript{L}~\cite{lin2004rouge} evaluation scripts. BLEU measures the average $n$-gram precision on a set of reference sentences, with a penalty for overly short sentences. BLEU-$n$ is BLEU score that uses up to $n$-grams for counting co-occurrences. METEOR is a recall-oriented metric, which calculates the similarity between generations and references by considering synonyms, stemming and paraphrases. ROUGE is commonly employed to evaluate $n$-grams recall of the summaries with gold-standard sentences as references. ROUGE\textsubscript{L} (measured based on longest common subsequence) results are reported.
  
% Results of automatic evaluations are displayed in Table~\ref{tab:results}.

\subsection{Human Evaluation}

We also perform human evaluation studies to measure the quality of questions generated by our system and the H\&S system. We consider two modalities: \emph{naturalness}, which indicates the grammaticality and fluency; and \emph{difficulty}, which measures the sentence-question syntactic divergence and the reasoning needed to answer the question. We randomly sampled 100 sentence-question pairs. We ask four professional English speakers to rate the pairs in terms of the modalities above on a 1--5 scale (5 for the best). We then ask the human raters to give a ranking of the questions according to the overall quality, with ties allowed.

% Use significance likelihood-ratio test for pairwise preference experiments?~\cite{serban2016factoid30m}. 
  
% % 10-fold cross-validation, evaluate on a held-out test set, with 9,200 snippet question pairs.
% avg. distance between snip. and predicted question, we want to encourage the literal similarity.

% \section{Results}

\begin{table}[t]
% \small
\centering
\resizebox{\columnwidth}{!}{%
    \begin{tabular}{lccccc} 
    \toprule
      & Naturalness & Difficulty & Best \% & Avg. rank\\\midrule
    % H\&S & 2.83 & 1.91 & 21.88 &2.28 \\
    % Ours & \textbf{3.39} & \textbf{2.97}\textsuperscript{*} &  \textbf{41.75}\textsuperscript{*} & \textbf{1.92}\textsuperscript{**} \\ 
    % Human & 3.71  & 2.61 & 61.96 & 1.53 \\ 
    
H\&S  & 2.95 & 1.94 & 20.20 & 2.29 \\
Ours  & \textbf{3.36} & \textbf{3.03}\textsuperscript{*} & \textbf{38.38}\textsuperscript{*} & \textbf{1.94}\textsuperscript{**} \\\midrule
Human & 3.91 & 2.63 & 66.42 & 1.46 \\
    \bottomrule
    % Meaningfulness & 0 & 0 \\ \bottomrule
    \end{tabular}}
  \caption{Human evaluation results for question generation. Naturalness and difficulty are rated on a 1--5 scale (5 for the best). Two-tailed t-test results are shown for our method compared to H\&S (statistical significance is indicated with $^{*}$($p$ < 0.005), $^{**}$($p$ < 0.001)).}
\label{tab:human}
\vspace{-0.2cm}
\end{table}
%   \NoteXinya{Use two-paired t-test for the \textit{average} of human scores, \emph{five scale point}.}

\section{Results and Analysis}
\label{sec:discussion}

\begin{figure}[!tb]
\begin{framed}
\small
  \noindent \textbf{Sentence 1}: the largest of these is the eldon square shopping centre , one of the largest city centre shopping complexes in the uk . 
  
  \vspace{0.02cm}\noindent \textbf{Human}: what is one of the largest city center shopping complexes in the uk ?
  
  \vspace{0.02cm} \noindent \textbf{H\&S}: what is the eldon square shopping centre one of ?
 
  \vspace{0.02cm} \noindent \textbf{Ours}: what is one of the largest city centers in the uk ?
  
  \par 
  \vspace{0.4cm} 
  \noindent \textbf{Sentence 2}: free oxygen first appeared in significant quantities during the paleoproterozoic eon -lrb- between 3.0 and 2.3 billion years ago -rrb- .  
  
  \vspace{0.02cm}\noindent \textbf{Human}: during which eon did free oxygen begin appearing in quantity ?
  
  \vspace{0.02cm} \noindent \textbf{H\&S}: what first appeared in significant quantities during the paleoproterozoic eon ?
  
  \vspace{0.02cm} \noindent \textbf{Ours}: how long ago did the paleoproterozoic exhibit ?
  
  \par 
  \vspace{0.4cm} 
  \noindent \textbf{Sentence 3}: inflammation is one of the first responses of the immune system to infection . 
  
  \vspace{0.02cm}\noindent \textbf{Human}: what is one of the first responses the immune system has to infection ?
  
  \vspace{0.02cm} \noindent \textbf{H\&S}: what is inflammation one of ?
  
  \vspace{0.02cm} \noindent \textbf{Ours}: what is one of the first objections of the immune system to infection ?
  
  \par 
  \vspace{0.4cm} 
  \noindent \textbf{Sentence 4}: tea , coffee , sisal , pyrethrum , corn , and wheat are grown in the fertile highlands , one of the most successful agricultural production regions in Africa.
  
  \vspace{0.02cm}\noindent \textbf{Human}: (1) where is the most successful agricultural prodcution regions ? (2) what is grown in the fertile highlands ?
  
  \vspace{0.02cm} \noindent \textbf{H\&S}: what are grown in the fertile highlands in africa ?
  
  \vspace{0.02cm} \noindent \textbf{Ours}: what are the most successful agricultural production regions in africa ?
  
  \par 
  \vspace{0.4cm} 
  \noindent \textbf{Sentence 5}: as an example , income inequality did fall in the united states during its high school movement from 1910 to 1940 and thereafter . 
  
  \vspace{0.02cm}\noindent \textbf{Human}: during what time period did income inequality decrease in the united states ?
  
  \vspace{0.02cm} \noindent \textbf{H\&S}: where did income inequality do fall during its high school movement from 1910 to 1940 and thereafter as an example ?
  
  \vspace{0.02cm} \noindent \textbf{Ours}: when did income inequality fall in the us ?
  
  \par 
  \vspace{0.4cm} 
  \noindent \textbf{Sentence 6}: however , the rainforest still managed to thrive during these glacial periods , allowing for the survival and evolution of a broad diversity of species .
  
  \vspace{0.02cm}\noindent \textbf{Human}: did the rainforest managed to thrive during the glacial periods ?
  
  \vspace{0.02cm} \noindent \textbf{H\&S}: what allowed for the survival and evolution of a broad diversity of species?
  
  \vspace{0.02cm} \noindent \textbf{Ours}: why do the birds still grow during glacial periods ?
 
  \par 
  \vspace{0.4cm} 
  \noindent \textbf{Sentence 7}: maududi founded the jamaat-e-islami party in 1941 and remained its leader until 1972.
  
  \vspace{0.02cm}\noindent \textbf{Human}: when did maududi found the jamaat-e-islami party ?

  \vspace{0.02cm} \noindent \textbf{H\&S}: who did maududi remain until 1972 ?
  
  \vspace{0.02cm} \noindent \textbf{Ours}: when was the jamaat-e-islami party founded ?

\end{framed}
\vspace{-0.5cm}
\caption{Sample output questions generated by human (ground truth questions), our system and the H\&S system. }
% Here \textbf{Human} means ground truth questions.}
\vspace{-0.4cm}
\label{fig:example}
\end{figure}

\begin{table*}[!htb]
\centering
\footnotesize
    \begin{tabular}{lp{1.5cm}p{.9cm}p{.9cm}p{.9cm}p{.9cm}p{.9cm}p{.9cm}p{.9cm}p{.9cm}p{.9cm}p{.9cm}} 
    \toprule
    \multirow{2}{*}{Category}  & \multirow{2}{*}{(\%)}  & \multicolumn{3}{c}{H\&S} & \multicolumn{3}{c}{Ours} & \multicolumn{3}{c}{Ours + paragraph} \\ \cmidrule(r{4pt}){3-5} \cmidrule(l){6-8} \cmidrule(l){9-11}
    & & {\scriptsize BLEU-3}& {\scriptsize BLEU-4}& {\scriptsize METEOR} & {\scriptsize BLEU-3}& {\scriptsize BLEU-4}& {\scriptsize METEOR} & {\scriptsize BLEU-3}&  {\scriptsize BLEU-4} & {\scriptsize METEOR} \\ \midrule
w/ sentence   & 70.23 (243) & 20.64 & 15.81 & 16.76 & \textbf{24.45} & \textbf{17.63} & 17.82 & 24.01 & 16.39 & \textbf{19.19} \\
w/ paragraph & 19.65 (68)  & 6.34  & < 0.01  & 10.74 & 3.76  & < 0.01  & 11.59 & \textbf{7.23} & \textbf{4.13} & \textbf{12.13} \\\midrule
% w/ article   & 0.58 (2)    & --     & --     & --     & --     & --     & --     & --     & --     & --     \\
% not askable  & 9.54 (33)   & -     & -     & -     & -     & -     & -     & -     & -     & -     \\\midrule
All\textsuperscript{*}          & 100 (346)   & 19.97 & 14.95 & 16.68 & 23.63 & \textbf{16.85} & 17.62 & \textbf{24.68} & 16.33 & \textbf{19.61} \\ \bottomrule
    \end{tabular}
  \caption{An estimate of categories of questions of the processed dataset and per-category performance comparison of the systems. The estimate is based on our analysis of the 346 pairs from the dev set. Categories are decided by the information needed to generate the question. Bold numbers represent the best performing method for a given metric. ${}^{*}$Here, we leave out performance results for ``w/ article'' category (2 samples, 0.58\%) and ``not askable'' category (33 samples, 9.54\%).}
\label{tab:category}
\vspace{-0.2cm}
\end{table*}

Table~\ref{tab:results} shows automatic metric evaluation results for our models and baselines. Our model which only encodes sentence-level information achieves the best performance across all metrics. We note that IR performs poorly, indicating that memorizing the training set is not enough for the task. The baseline DirectIn performs pretty well on BLEU and METEOR, which is reasonable given the overlap statistics between the sentences and the questions (Figure~\ref{fig:overlap}). H\&S system's performance is on a par with DirectIn's, as it basically performs syntactic change without paraphrasing, and the overlap rate is also \linepenalty=1000 high.

Looking at the performance of our three models, it's clear that adding the pre-trained embeddings generally helps. While encoding the paragraph causes the performance to drop a little, this makes sense because, apart from useful information, the paragraph also contains much noise.

Table~\ref{tab:human} shows the results of the human evaluation. We see that our system outperforms H\&S in all modalities. Our system is ranked best in 38.4\% of the evaluations, with an average ranking of 1.94. An inter-rater agreement of Krippendorff's Alpha of 0.236 is achieved for the overall ranking. The results imply that our model can generate questions of better quality than the H\&S system. An interesting phenomenon here is that human raters gave higher score for our system's outputs than the human questions. One potential explanation for this is that our system is trained on {\em all} sentence-question pairs for one input sentence, while we randomly select one question among the several questions of one sentence as the human generated question, for the purpose of rating. Thus our system's predictions tend to be more diverse.

\begin{figure}[t]
\centering
\small
\includegraphics[scale=.48]{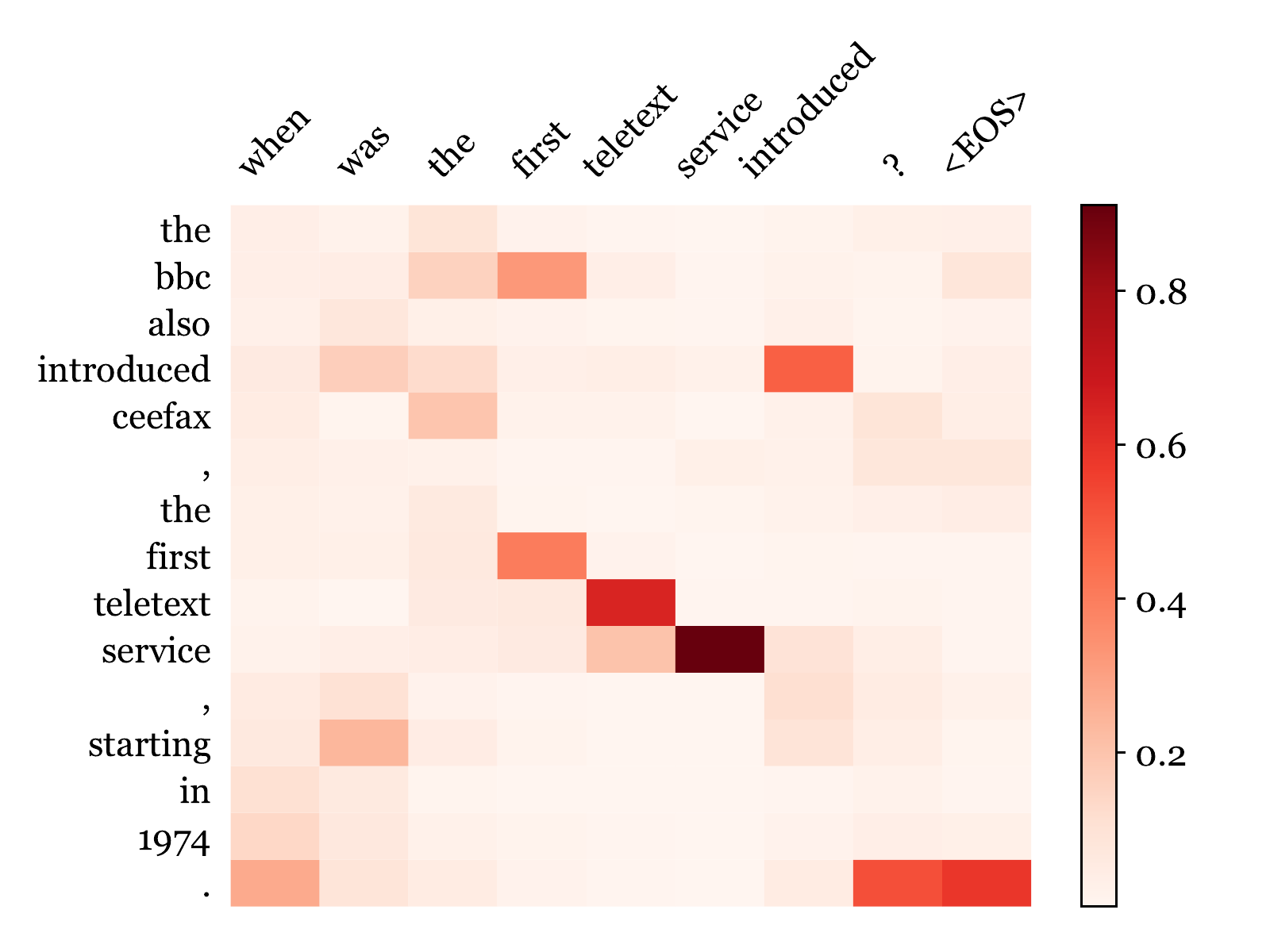}
\vspace{-0.5cm}
\caption{Heatmap of the attention weight matrix, which shows the soft alignment between the sentence (left) and the generated question (top).}
\label{fig:heatmap}
\vspace{-0.1cm}
\end{figure}

For our qualitative analysis, we examine the sample outputs and the visualization of the alignment between the input and the output. In Figure~\ref{fig:example}, we present sample questions generated by H\&S and our best model. We see a large gap between our results and H\&S's. For example, in the first sample, in which the focus should be put on ``the largest.'' Our model successfully captures this information, while H\&S only performs some syntactic transformation over the input without paraphrasing. However, outputs from our system are not always ``perfect'', for example, in pair 6, our system generates a question about the reason why birds still grow, but the \emph{most related} question would be why many species still grow. But from a different perspective, our question is more challenging (readers need to understand that birds are one kind of species), which supports our system's performance listed in human evaluations (See Table~\ref{tab:human}). It would be interesting to further investigate how to interpret why certain irrelavant words are generated in the question. Figure~\ref{fig:heatmap} shows the attention weights ($\alpha_{i,t}$) for the input sentence when generating each token in the question. We see that the key words in the output (``introduced'', ``teletext'', etc.) aligns well with those in the input sentence.

Finally, we do a dataset analysis and fine-grained system performance analysis. We randomly sampled 346 sentence-question pairs from the dev set and label each pair with a category.~\footnote{The IDs of the questions examined will be made available at~\url{https://github.com/xinyadu/nqg/blob/master/examined-question-ids.txt}.} The four categories are determined by \emph{how much} information is needed to ask the question. To be specific, ``w/ sentence'' means it only requires the sentence to ask the question; ``w/ paragraph'' means it takes other information in the paragraph to ask the question; ``w/ article'' is similar to ``w/ paragraph''; and ``not askable'' means that world knowledge is needed to ask the question or there is mismatch of sentence and question caused by annotation error.

 Table~\ref{tab:category} shows the per-category performance of the systems. Our model which encodes paragraph information achieves the best performance on the questions of ``w/ paragraph'' category. This verifies the effectiveness of our paragraph-level model on the questions concerning information outside the sentence.

\section{Conclusion and Future Work}

We have presented a fully data-driven neural networks approach to automatic question generation for reading comprehension. We use an attention-based neural networks approach for the task and investigate the effect of encoding sentence- vs. paragraph-level information. Our best model achieves state-of-the-art performance in both automatic evaluations and human evaluations.

Here we point out several interesting future research directions. Currently, our paragraph-level model does not achieve best performance across all categories of questions. We would like to explore how to better use the paragraph-level information to improve the performance of QG system regarding questions of all categories. Besides this, it would also be interesting to consider to incorporate mechanisms for other language generation tasks (\eg, copy mechanism for dialogue generation) in our model to further improve the quality of generated questions.

% Considering the difference between our results and the human generated questions, as well as the manual dataset analysis, we believe that the task is promising, and hope to draw the attention of the community to it.

\section*{Acknowledgments}

We thank the anonymous ACL reviewers, Kai Sun and Yao Cheng for their helpful suggestions. We thank Victoria Litvinova for her careful proofreading. We also thank Xanda Schofield, Wil Thomason, Hubert Lin and Junxian He for doing the human evaluations.

% include your own bib file like this:
%\bibliographystyle{acl}
%\bibliography{acl2017}

% \clearpage

\bibliography{acl2017}
\bibliographystyle{acl_natbib}

% \appendix

% \section{Supplemental Material}
% \label{sec:supplemental}

\end{document}